# QUALITY ASSESSMENT OF IMAGE MATCHERS FOR DSM GENERATION - A COMPARATIVE STUDY BASED ON UAV IMAGES

Rongjun QIN[1], Prof. em. Armin GRUEN[2], Prof. Clive FRASER[3]

[1] PhD student at Future Cities Laboratory, Singapore-ETH Center, 1 CREATE Way #06-01 CREATE Tower
Singapore 138602
Tel: +65-8389-5276
Email: rqin@student.ethz.ch

[2] Principal Investigator of the Future Cities Laboratory, Chair of Information Architecture
ETH Zuerich, Wolfgang-Pauli-Strasse 27, CH-8093 Zuerich Switzerland
Tel: +41-44633-3038
Email: agruen@geod.baug.ethz.ch

[3] Professorial Fellow, Dept. of Infrastructure Engineering, University of Melbourne, Vic 3010, Australia
Tel: +61-3-8344-4117
Email: c.fraser@unimelb.edu.au

**ABSTRACT:** Recently developed automatic dense image matching algorithms are now being implemented for DSM/DTM production, with their pixel-level surface generation capability offering the prospect of partially alleviating the need for manual and semi-automatic stereoscopic measurements. In this paper, five commercial/public software packages for 3D surface generation are evaluated, using 5cm GSD imagery recorded from a UAV. Generated surface models are assessed against point clouds generated from mobile LiDAR and manual stereoscopic measurements. The software packages considered are APS, MICMAC, SURE, Pix4UAV and an SGM implementation from DLR.

The evaluation is conducted on a typical urban scene of 354 m × 185 m size, containing buildings, roads, variable terrain and tropical vegetation. DSMs are initially generated by the five software packages and then co-registered to the ground reference data using least-squares 3D surface matching, which minimizes the squared sum of the Euclidean differences between the matched DSM and the ground reference. The RMSEs, standard deviations and the error distributions (the analysis of blunders in particular) are used for the evaluation of the matchers on both solid (buildings, road surfaces, bare earth, etc.) and 'soft' objects (trees, bushes, etc.). The analysis covers the full dataset, the case of solid objects only, and finally a buildings-only DSM. The results of the experiments provide a useful indicator of matcher performance under the particular UAV imagery conditions considered.

## 1. INTRODCUTION

Image matching is an important component in the process of automated DSM (Digital Surface Model) generation and 3D modeling. Over the past 15 years or so, airborne LiDAR (Light Detection and Ranging) data has been seen as a preferred reliable source for accurate DSM generation, as well as in orthophoto production. However, its limited flexibility, high cost and high dependence on accurate navigation sensors limit its application with UAVs (unmanned aerial vehicles). In recent years, the development of Dense Image Matching (DIM) has provided renewed impetus to image-based DSM generation and 3D urban surface modeling, and the potential of DIM for alleviating the manual work in traditional stereo photogrammetric measurement procedures has been recognized (Gehrke et al., 2010). Indeed, photogrammetry is now widely considered as the optimal approach for highest-resolution 3D city model generation. DIM is equally well suited to imagery from professional or amateur cameras, and in both cases yields surface models with controlled accuracy. Moreover, given the high redundancy afforded by multi-overlap imagery, there remain opportunities for further developments in the design of matching algorithms and thus DIM will likely continue to be a hot research topic for some time (Gruen, 2012).

There are presently a number of open-source and commercial DIM software packages available, these having emanated from both the computer vision and photogrammetry communities. It is unsurprising that there are variations in the performance of these DIM implementations, which we will refer to here as 'matchers', especially given that their range of application covers satellite, aerial and terrestrial imagery, both vertical/normal and oblique/convergent. Indeed it is fair to say that there is presently no 'leading' DIM software suite within the commercial market place. However, amongst the large number of algorithms developed for dense matching

(Scharstein and Szeliski, 2002), there is a 'leading' algorithm, that being Semi-Global Matching (SGM) (Hirschmüller, 2008). Whereas there has been recent benchmark testing of DIM (e.g. Haala, 2013), this has mainly focused upon medium scale topographic surface modeling. There has been limited assessment to date of the performance of different matchers when applied to high-resolution UAV imagery covering dense urban environments. This paper addresses this topic, with the focus being upon matching performance in practical situations.

The evaluation of a matcher is usually performed by comparing the results to a reference data set of higher accuracy (Remondino et al., 2014), the reference data usually comprising LiDAR point clouds or reconstructed 3D models or DSMs with reliable point measurement (usually generated by manual or semi-automated approaches). In this work, we evaluate the DSM generation accuracy, through a comparison to a reference 3D model, of a sample of open-source and commercial matchers applied to UAV vertical images recorded with a consumer-grade camera. The reference 3D surface model, manually and semi-automatically generated, is compared to the DIM generated 3D point clouds using the method of least-squares surface matching, via the LS3D software system (Akca, 2010). LS3D computes an optimal registration through a minimization of the Euclidean distance between two surface models. This was already used to check the quality of produced 3D models in an Ordnance Survey project (Akca et al., 2010). The DIM software packages evaluated are APS (www.menci.com), MICMAC (Deseilligny and Clery, 2011), SURE (nframes.com), Pix4uav (pix4d.com) and an SGM implementation from DLR in Germany (Hirschmüller, 2008), which will be referred to simply as DLR.

Due to the limitation of lightweight navigation systems for UAV-based LiDAR, photogrammetric methods currently offer the only practical means for UAV-based high accuracy mapping. Thus, the reported quantitative analysis of the selected currently available matchers is intended to provide a useful insight into their potential for accurate, high-resolution DSM and 3D city model surface generation from UAV imagery. When considering this problem one should be aware that, in case of very complex scenes, as given here, the higher the data resolution (in our case 5 cm pixel footprint) the more difficult becomes the image matching and also the procedure of quality control.

## 2. DATA PREPARATION

### 2.1. Reference data

*2.1.1 3D model generation*

The reference data used for the evaluation was produced in an earlier 3D modeling project (Gruen et al., 2013; Qin et al., 2012), which aimed to generate a detailed 3D model of the National University of Singapore (NUS) campus area. In that project, 857 images with a GSD of 5 cm were taken from a height of 150 meters above ground, and the images were geo-referenced with accurate ground control points (GCP) from GPS-based ground survey. The triangulation accuracy of the reference imagery was slightly above one pixel (since natural targets were used as GCPs). At roughly the same time, LiDAR point clouds were acquired with a RIEGL WMX-250 system in terrestrial mobile mapping mode. In total, the images cover an area of 2.2 km$^2$ and the LiDAR point clouds cover 16 km of the main roads in the mapping area. The point clouds were geo-referenced with 169 control points derived from the UAV images, leading to an accurate co-registration of both data.

3D building models were then generated semi-automatically with the CC-modeler software (Gruen and Wang, 1998), with key points measured in stereo mode. The 3D terrain data mainly comprised: 1) manually measured feature points and lines, 2) points and features extracted from the LiDAR point clouds, and 3) survey points under the tree canopy. Trees were modeled using a combination of generic and reality-based modeling, where tree tops and radii were measured from UAV images, and fitted with pre-selected models (Gruen et al., 2013).

*2.1.2. Test area*
This evaluation utilized a small sub-area of 354 m × 185 m of the overall 3D model. The land cover in the selected area, which lies in the middle of the larger mapping area, comprised buildings, roads and trees (see Figure 1). A 5 × 5 image block was considered for the comparative evaluation of the different matchers. In order to assess the matcher performance more comprehensively, tree canopies in the test area were manually measured with profile points and break lines. The previously modeled trees were not used.

The reference 3D model then comprised individual models of buildings, roads, trees and the terrain.
For the preparation of the reference dataset a particular implementation aspect of the LS3D (Least Squares 3D Matching) software had to be considered. LS3D does not accept mixed data formats, as they were given in the dataset: man-made objects in v3d format of CC-Modeler and the other points (terrain, trees, etc.) as wireframe.

Therefore, a point cloud was derived directly from the 3D meshes making up the reference data using a Z-buffer algorithm (Sutherland et al., 1974) to obtain the upper layer of the DSM. This resulted in a total of 6,436,449 surface points (see Figure 1 right).

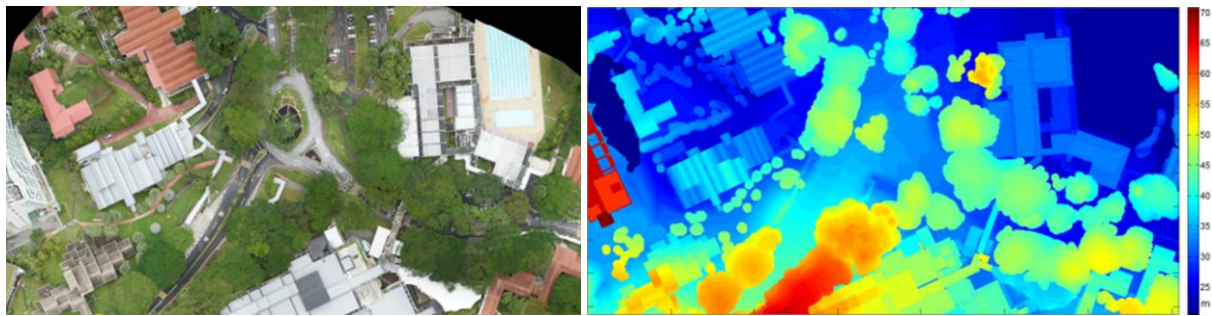

Figure 1. Orthophoto of the test area (left) and the Z component of the reference 3D model (right).

**2.2 DIM point cloud generation and pre-processing**

In principle, the evaluation of the different matchers should be based on common orientation parameters for the images, which helps to concentrate the evaluation on the image matching process itself. However, this can sometimes be difficult to achieve since commercial matchers, and even some open-source matchers, invariably only accept their own format of orientation and camera calibration. Some DIM software packages do not even support the import of exterior orientation parameters. For this reason, point clouds from the selected matchers were first generated via their respective standard processing pipelines and then co-registered to the reference data in object space.

Of the five matchers/DIM software evaluated, namely APS, DLR, MICMAC, Pix4uav and SURE, only SURE could readily adopt the same orientation parameters as the previous 3D modeling project. DLR utilized the geo-referencing results from Leica's LPS software, and the $5 \times 5$ image blocks were geo-referenced by the remaining three matchers using their own orientation modules (MICMAC adopted parameters computed by Apero) with 5 GCPs used for the process. For the multi-stereo algorithms, only points with at least three imaging rays (i.e. object points seen in at least two stereo models) were retained in the generated point clouds.

It should be noted that a multi-stereo matching algorithm (as an approximation of strict multi-image matching) usually generates a point cloud file per image, and parts of the objects/ground underneath an upper canopy might be visible for some of the stereo pairs. This leads to multi-layer point clouds. Therefore a Z-buffer strategy was adopted to obtain the highest layer of the point clouds. A raster grid with a GSD of 10 cm was first pre-allocated, and then for each cell the highest of the included Z values was adopted. In order to keep the original measurements, the Z-buffer process not only stored the adopted Z value for each cell, but also its X, Y coordinates so as to ensure that the original measurements from the matchers were being compared.

To distinguish the matching performance for soft (i.e. trees) and hard objects (impervious surfaces), a separate evaluation was undertaken for three different object classes: 1) for the whole DSM, 2) for the DSM without trees and 3) for buildings only. The point cloud masking for the latter two classes was carried out manually. The building model actually only consists of roofs, but includes also sheltered pedestrian walkways and other man-made objects.

**3. LEAST-SQUARES 3D SURFACE MATCHING**

As discussed in Section 2, due to software limitations, the point cloud generation for different software packages could not be based on the same set of orientation parameters. A co-registration process using the LS3D method developed by (Akca, 2010, Gruen and Akca, 2005) was therefore employed. LS3D is a rigorous algorithm that minimizes the squared sum of the Euclidean differences between the matched point clouds and the reference DSM. It uses a generalized Gauss-Markov formulation to estimate the 7-parameter similarity transformation (X, Y, Z, ω, φ, κ and scale) between overlapping 3D surfaces.

The LS3D co-registration process serves two purposes. Firstly, it is used to eliminate systematic errors between the matched point clouds and the reference data, as well as gross errors from matching artifacts (if desired) and, secondly, it gives accuracy values that describe the quality of the fit of both surfaces. LS3D is an iterative process, and for each iteration the procedure performs a blunder check procedure, namely there is an elimination of correspondences that have higher residuals than $K\delta$, where $\delta$ is the standard deviation (STD) of the Euclidean distance in the current iteration, and $K$ is a user-defined value. $K$ should be neither too small (e.g. less than 4) nor too big (e.g. greater than 10), as a small value may eliminate correct matches and a large value may cause the computation to be distorted by incorrect correspondences. In our case however we wanted to keep the blunders in the system, because they are the result of the matchers' performance. Therefore the value of $K = 100$ was adopted as an empirical estimate for this project.

## 4. MATCHER EVALUTION

Based on the LS3D method, a two-step strategy for the matcher evaluation was adopted, noting that both the reference data and DIM output were in the form of point clouds:

1) In an initial LS3D co-registration run only six parameters (X, Y, Z, ω, φ, and κ) were estimated, in recognition of there being no scale difference between the matcher point clouds and the reference data. This stage also incorporated blunder detection with $K = 5$.

2) A second run of LS3D was then performed, this time with only one iteration with very high $K$ (i.e. 100) to include blunders. The transformation parameters from the initial run were held fixed, as the purpose here was simply to compute the final Euclidean distances (discrepancies) between the two datasets without blunders being eliminated.

As mentioned, the two-step LS3D process was performed on the whole DSM, on the DSM without trees, and on building objects. This was in recognition of the fact that the matching quality for different objects varies. For example, the results for tree canopies may not be so important due to natural seasonal variations, and those for buildings might assume more importance, since there is a direct impact on the quality of the resulting 3D building models. Therefore, it is important to understand the performance of a matcher under different scenarios.

### 4.1. Statistical summary of results

Tables 1-3 show accuracy statistics determined in the evaluation of each matcher, for each of the three reference DSM cases. The RMSE (Root Mean Square Error) indicates the Euclidean distance between the two datasets, while STD gives a bias-free measure of the standard deviation. The Matching Percentage denotes the portion of the matcher data where blunder-free correspondences were found between the reference and matcher point clouds. That is, the difference to 100% indicates the percentage of blunders, based on $K = 5$. Therefore, the Matching Percentage value can be used to indicate the percentage of correct 3D point measurements. Finally, Completeness denotes the portion of the matcher data where valid correspondences from the reference DSM were found, inclusive of blunders. The missing correspondences result from points whose partners fall outside the reference frame.

Table 1. Evaluation results for the whole DSM

|  | APS | DLR | MICMAC | PIX4D | SURE |
|---|---|---|---|---|---|
| Matching percentage (%) | 72.9 | 83.4 | 78.7 | 80.2 | 84.7 |
| Completeness (%) | 95.3 | 97.9 | 97.8 | 98.0 | 97.7 |
| $\delta_0$ (m) | 0.29 | 0.29 | 0.31 | 0.26 | 0.26 |
| RMSE (m) | 0.79 | 0.55 | 0.62 | 0.63 | 0.54 |
| STD (m) | 0.78 | 0.55 | 0.61 | 0.63 | 0.54 |
| MEAN (m) | -0.10 | -0.04 | -0.07 | -0.06 | -0.03 |
| Minimum (m) | -6.42 | -6.28 | -5.96 | -6.14 | -5.96 |
| Maximum (m) | 6.18 | 5.44 | 5.50 | 5.42 | 5.34 |

Table 2. Evaluation results for the DSM without trees

|  | APS | DLR | MICMAC | PIX4D | SURE |
|---|---|---|---|---|---|
| Matching percentage (%) | 71.5 | 83.5 | 83.1 | 79.2 | 85.4 |
| Completeness (%) | 90.7 | 94.6 | 95.2 | 93.2 | 94.8 |
| $\delta_0$ (m) | 0.25 | 0.25 | 0.23 | 0.22 | 0.20 |
| RMSE (m) | 0.75 | 0.47 | 0.54 | 0.58 | 0.46 |
| STD (m) | 0.74 | 0.47 | 0.53 | 0.57 | 0.46 |
| MEAN (m) | -0.15 | -0.04 | -0.09 | -0.07 | -0.03 |
| Minimum (m) | -5.89 | -6.15 | -6.19 | -6.03 | -5.90 |
| Maximum (m) | 5.51 | 4.57 | 3.72 | 5.02 | 4.05 |

Table 3. Evaluation results for the DSM of buildings only

|  | APS | DLR | MICMAC | PIX4D | SURE |
|---|---|---|---|---|---|
| Matching percentage (%) | 75.7 | 84.3 | 84.8 | 79.9 | 85.2 |
| Completeness (%) | 84.6 | 88.0 | 89.9 | 85.9 | 87.8 |
| $\delta_0$ (m) | 0.23 | 0.20 | 0.17 | 0.19 | 0.17 |
| RMSE (m) | 0.51 | 0.28 | 0.34 | 0.43 | 0.24 |
| STD (m) | 0.50 | 0.28 | 0.33 | 0.42 | 0.24 |
| MEAN (m) | -0.08 | -0.02 | -0.04 | -0.07 | -0.02 |
| Minimum (m) | -5.90 | -5.56 | -5.20 | -5.80 | -5.47 |
| Maximum (m) | 5.48 | 4.94 | 4.05 | 4.76 | 4.41 |

The $\delta_0$ value represents the standard deviation of the co-registration process of step 1, which effectively indicates the RMS Euclidean distance between the matching results and the reference data after blunder elimination. It can be seen that in all cases SURE has the smallest $\delta_0$. All matchers derive more than 70 percent "good" correspondences in comparison with the reference data. As expected, the matchers have larger RMSEs when the whole DSM is considered, and have the smallest RMSEs when using only building roofs for comparison.

From Tables 1-3 it can be seen that SURE yields the smallest RMSEs (0.54, 0.46 and 0.24) and also the smallest $\delta_0$ in all three scenarios. This means that it performs best in all versions with and without blunders. Under the particular conditions of the evaluation, SURE displayed the highest measurement accuracy amongst those tested and also the best matching percentage. However, the differences to the other matchers are not very large.

**4.2. Error Distributions**

The estimates of RMSE and STD are key factors for assessing the matching performance, but it is equally important to analyze the error distributions, since this provides information about DIM reliability (e.g. size and number of blunders). A histogram of the surface discrepancies from different matchers for the case of the whole DSM is shown in Figure 2. It can be seen that the PIX4D matcher has the highest zero-centered peak, while it also has a large number of blunders (see its tails, larger than 0.4 or smaller than -0.4 and the blunder percentage of Table 4).

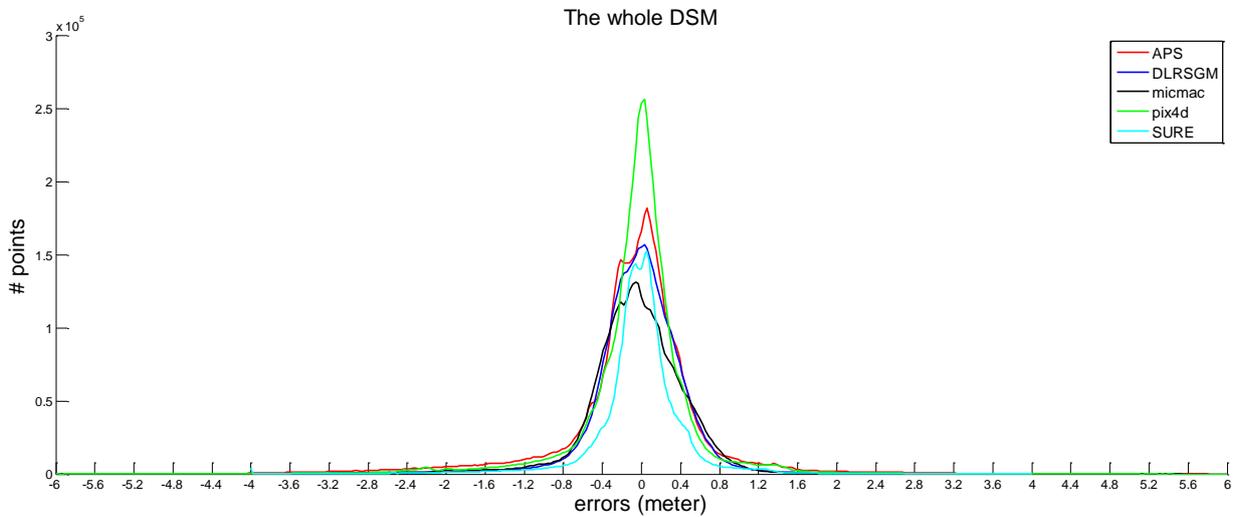

Figure 2. Error distribution of different matching results for the whole DSM.

Table 4. Number of blunders larger than 3 times STD

| | APS | | DLR | | MICMAC | | PIX4D | | SURE | |
|---|---|---|---|---|---|---|---|---|---|---|
| | number | percentage | number | percentage | number | percentage | number | percentage | number | percentage |
| The whole DSM | 150317 | 2.88 | 97903 | 2.19 | 92077 | 2.23 | 147294 | 2.77 | 106126 | 2.34 |
| DSM without trees | 99622 | 3.16 | 53072 | 1.91 | 63064 | 2.49 | 88963 | 2.72 | 66139 | 2.33 |
| Buildings only | 31045 | 2.32 | 14458 | 1.19 | 26162 | 1.98 | 41336 | 2.96 | 16847 | 1.31 |

All matchers show their best results in the buildings-only scenario (Table 3). The DLR matcher reveals the smallest percentage number of blunders (Table 4), but the absolute number of blunders is still very high (up to 150,000 for APS and the complete model and still 40,000 for PIX4D in case of buildings only).
Figure 3 shows the Euclidean distances between the reference 3D data and the point cloud generated by SURE. Most of the large blunders (>0.8 m) occur in areas of vegetation and on some building edges. Obviously the swimming pool (red area in the upper right corner) also generates blunders, which was to be expected. Fairly large gaps indicating failed matches can be seen in the test area. For all matchers, these being mainly in areas covered by vegetation, as well as on some roof tops that display homogeneous texture.

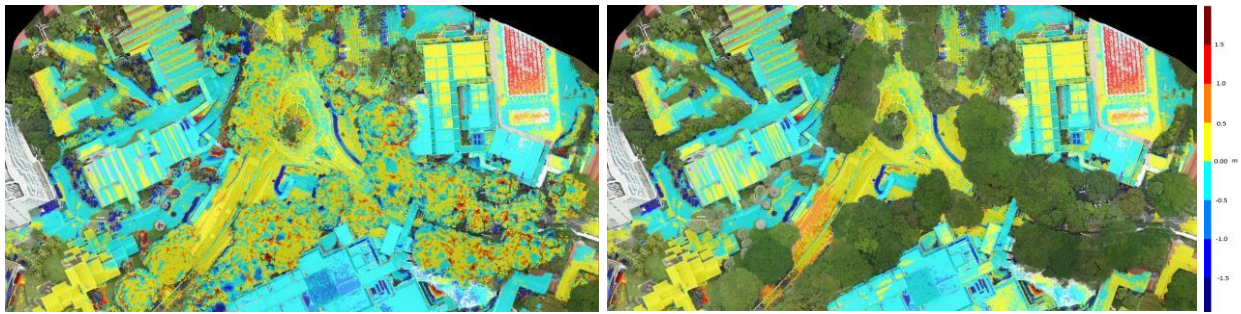

Figure 3. The error distribution of SURE results overlaid on the orthophoto. Left: the whole DSM, right: the DSM without trees.

## 5. CONCLUSIONS

From the evaluation of five different DIM software packages applied to 5cm GSD UAV imagery, useful insights were gained by investigating the discrepancies between the point clouds generated by each of the matchers and a reference 3D surface model. The LS3D method was used to accurately register the matcher DSMs to the reference data, and the Euclidean distances between the co-registered DSMs was used as a basis for the accuracy evaluation covering three different scenarios: the overall DSM, the DSM with vegetation masked out, and the buildings-only DSM.

All the matchers produced their best accuracy in the buildings-only case, with less accurate results being obtained over 'soft' vegetation canopies. Of the five matchers evaluated, SURE yielded the best accuracy (minimum RMSE and $\delta_0$), although the differences to the other matchers are not very large. Actually, the accuracy performance of the five matchers differed from each other by less than was expected.

The overall accuracy level achieved in the DSM generation was not particularly high, with RMSEs being between 5 and 11 pixels with SURE. Even if blunders are taken out (based on $K = 5$), the accuracy still is at the 3 to 5 pixel level. This illustrates that DIM limitations are most likely to be seen with very high resolution imagery (pixel size 5 cm in this case). A persistent problem common to all the matchers considered was the relatively large number of blunders and gaps (lack of measurements) in the generated point clouds, these being particularly prevalent in vegetated areas.
Here a basic problem with very high resolution data quality control must be mentioned. In this case to generate high quality reference data is quite demanding. In our test we made no extra effort to generate this data, but we took it from the 3D campus model that was produced earlier with manual UAV image measurements and mobile laser-scan data. So for instance there are many small objects on the roofs of buildings, like installations for air conditioning, etc., which are not a permanent part of the architecture and as such have not been measured. Therefore these missing objects may also produce larger errors. In any case it is advised to do more tests of this kind before final conclusions can be drawn.

## 6. ACKNOWLEDGEMENT

This work was performed at the Singapore-ETH Centre for Global Environmental Sustainability (SEC), co-funded by the Singapore National Research Foundation (NRF) and ETH Zurich. The authors would like to thank Dr. Devrim Akca for providing the LS3D software and Menci software for providing evaluation license of APS.